\documentclass[conference]{IEEEtran}
\usepackage{cite}
\usepackage{amsmath,amssymb,amsfonts}
\usepackage{algorithmic}
\usepackage{graphicx}
\usepackage{textcomp}
\usepackage{xcolor}
\def\BibTeX{{\rm B\kern-.05em{\sc i\kern-.025em b}\kern-.08em
    T\kern-.1667em\lower.7ex\hbox{E}\kern-.125emX}}

\usepackage{hyperref}
\hypersetup{
    colorlinks=true,
    linkcolor=blue,
    filecolor=magenta,      
    urlcolor=cyan,
    pdftitle={Overleaf Example},
    pdfpagemode=FullScreen,
    }

\begin{document}

\title{FINNger - Applying artificial intelligence to ease math learning for children}

\author{\IEEEauthorblockN{Audibert, Rafael B.}
\IEEEauthorblockA{\textit{INF - UFRGS} \\
Porto Alegre, Brazil \\
rbaudibert@inf.ufrgs.br}
\and
\IEEEauthorblockN{Maschio, Vinícius M.}
\IEEEauthorblockA{\textit{INF - UFRGS} \\
Porto Alegre, Brazil \\
vmmaschio@inf.ufrgs.br}
}

\maketitle

\begin{abstract}
    Kids have an amazing capacity to use modern electronic devices such as tablets, smartphones, etc. This has been incredibly boosted by the ease of access of these devices given the expansion of such devices through the world, reaching even third world countries. Also, it is well known that children tend to have difficulty learning some subjects at pre-school. We as a society focus extensively on alphabetization, but in the end, children end up having differences in another essential area: Mathematics. With this work, we create the basis for an intuitive application that could join the fact that children have a lot of ease when using such technological applications, trying to shrink the gap between a fun and enjoyable activity with something that will improve the children knowledge and ability to understand concepts when in a low age, by using a novel convolutional neural network to achieve so, named FINNger.
\end{abstract}

\section{Introduction}
Kids have an amazing capacity to use modern electronic devices such as tablets, smartphones, etc. This has been incredibly boosted by the ease of access of these devices given the expansion of such devices through the world, reaching even third world countries\cite{b1}. So, an easy-to-follow step would be to use this technology when teaching/developing our children.
It is well known that children tend to have difficulty learning some subjects at pre-school. We as a society focus extensively on alphabetization, but in the end, children end up having differences in another essential area: Mathematics. We then focus on creating an intuitive application that could join the fact that children have a lot of ease when using such technological applications, trying to shrink the gap between a fun and enjoyable activity with something that will improve the children knowledge and ability to understand concepts when in a low age.

\section{Our Application}

\begin{figure}[htbp]
\centerline{\includegraphics[width=55mm,scale=0.5]{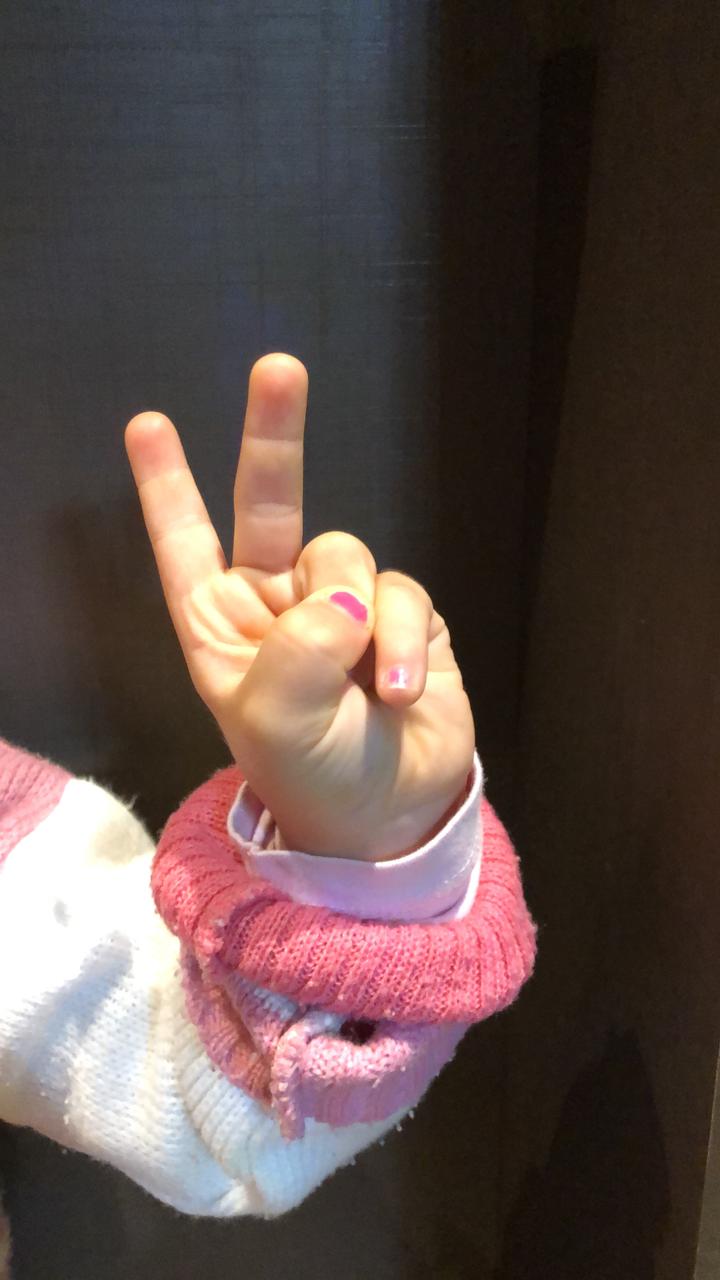}}
\caption{Example of a kid hand which we should label as $2$ raised fingers}
\label{kid_hand}
\end{figure}

On this assignment, we propose to build a desktop application where kids could, using their camera and/or any image capturing device, show to the computer their hands (see Fig. \ref{kid_hand}), and the computer would be able to infer how many fingers that child raised and be able to compute some basic arithmetic with it. Ideally, and in future work, this application should be able to run in a smartphone or a tablet, preferentially a low-end one, easing the access to non-developed countries (the ones which would benefit the most from it) to the application.

We propose two different approaches for this:

\begin{itemize}
    \item using an edge detection algorithm with a custom heuristic to compute how many raised fingers there is. This is mostly a baseline method.
    \item using a convolutional neural network trained in a big number of images so that the model can generalize enough so that it can differentiate between different hand sizes and colors, different backgrounds, and different illumination setups.
\end{itemize}

\section {Dataset}

To develop our application, a dataset was needed: both to test the edge detection algorithm and heuristic, and to train the neural network.

\subsection{koryakinp/fingers}
\begin{figure}
\begin{tabular}{cc}
{\includegraphics[width = 1.5in]{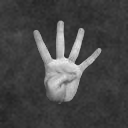}} &
{\includegraphics[width = 1.5in]{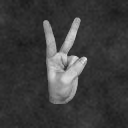}} \\
{\includegraphics[width = 1.5in]{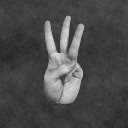}} &
{\includegraphics[width = 1.5in]{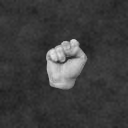}}
\end{tabular}
\caption{Example images available on \cite{b7} dataset}
\label{koryakinp_fingers}
\end{figure}

\cite{b7} is a dataset developed by Pavel Koryakin made available on \href{Kaggle}{https://kaggle.com}  with a public domain license. It consists of $21600$ black and white $128x128$ images of right and left hands, holding from $0$ to $5$ raised fingers, in a noisy background. This $21600$ images are  furter split on $18000$ images which will be used to train the dataset, and $3600$ which will be used to validate/test it. Some example of the available images can be seen on Fig. \ref{koryakinp_fingers}.

\subsection{Custom dataset}
However, we found that the above dataset wouldn't closely relate to reality since the images have a very specific illumination and pose, hardly related to a real-life image. Given that, we proposed to create a custom dataset where we could have images with different lighting and real-life background setups.

Therefore, we created a dataset with $3200$ images ($200$ for each number quantity on each hand), from which we used $2840$ ($85\%$) for the training dataset and the remaining $360$ for the testing/validation phase.

To generate it, we developed a custom application to make the task easier, being able to generate the full dataset in only 15 minutes, which provide us the ability to easily expand it to a bigger size in the future. The application (developed in Python) asks for the number of images that we want to generate, as well as which hand and quantity of numbers we are taking photos (so that we can automatically label it), and proceeds to take photos using a webcam.

\begin{figure}
\begin{tabular}{cc}
{\includegraphics[width = 1.5in]{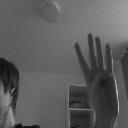}} &
{\includegraphics[width = 1.5in]{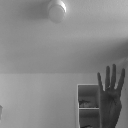}} \\
{\includegraphics[width = 1.5in]{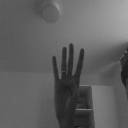}} &
{\includegraphics[width = 1.5in]{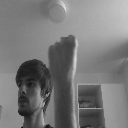}}
\end{tabular}
\caption{Example images from the custom dataset}
\label{custom_fingers}
\end{figure}

Some examples of pictures taken for this dataset can be seen on Fig. \ref{custom_fingers}. One can notice that the image even has the presence of the body of the person, and not only its hand, more closely relating to a real-life situation. 

\section{Implementation}

\subsection{Edge Detection}
Edge detection is a process that attempts to distinguish significant intensity changes from an image. This differentiation becomes relevant by the use of a regularizing filter operation. This process is done to being able to create a delimitation between the hand and the image's background.

\subsubsection{Distinguishing from background}
\begin{figure}[htbp]
\centerline{\includegraphics[width=50mm, scale=0.5]{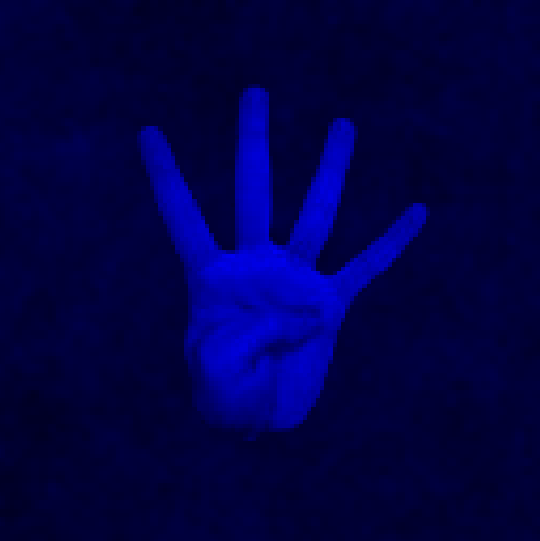}}
\caption{Image converted to HSV colors}
\label{HSV_hand}
\end{figure}

In the process to differentiate between the object we are trying to evaluate and the background, we start by converting the image colors to HSV using \cite{cv2}. With this result, we can separate the object within the background by the range of the colors, using a lower - RGB(0, 0, 255) - and a higher - RGB(0, 0, 90) - color skin for a range of the hand, as can be found on Fig. \ref{HSV_hand}.

\subsubsection{Separation}
\begin{figure}[htbp]
\centerline{\includegraphics[width=50mm,scale=0.5]{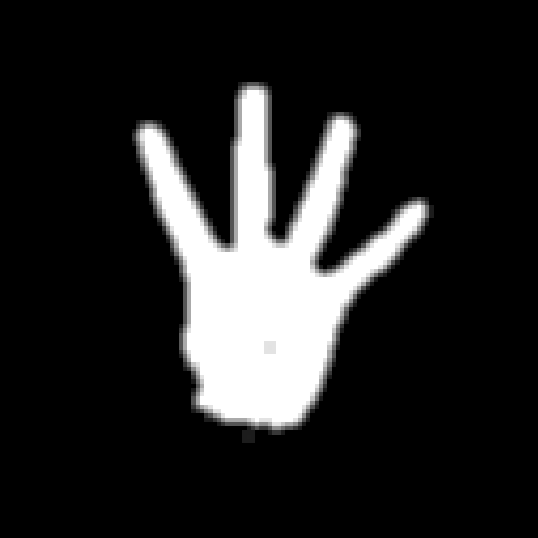}}
\caption{Mask with gaussian blur generated from Fig. \ref{HSV_hand}}
\label{mask_hand}
\end{figure}

With the color range and the HSV image found on Fig. \ref{HSV_hand}, we can create a mask of the object we need to distinguish. Using \cite{cv2} for the process, we generate a mask and apply on it a gaussian blur to remove unwanted noises, such as blank points, which can be found on Fig. \ref{mask_hand}.

\subsubsection{Distinguishing fingers}
\begin{figure}[htbp]
\centerline{\includegraphics[width=50mm,scale=0.5]{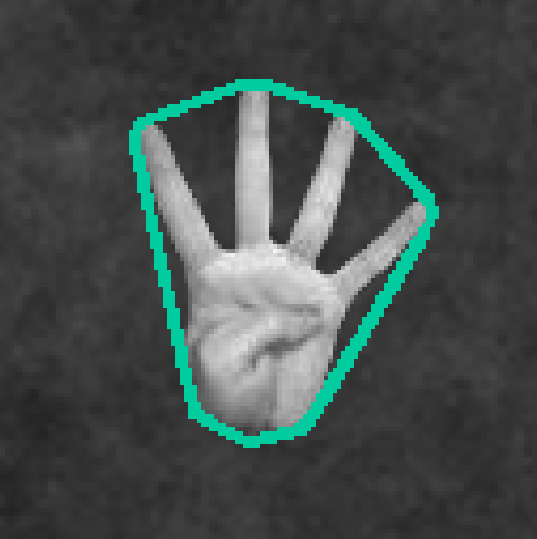}}
\caption{Contoured Image using Convex Hull from \cite{cv2}}
\label{contour_hand}
\end{figure}

\begin{figure}[htbp]
\centerline{\includegraphics[width=50mm,scale=0.5]{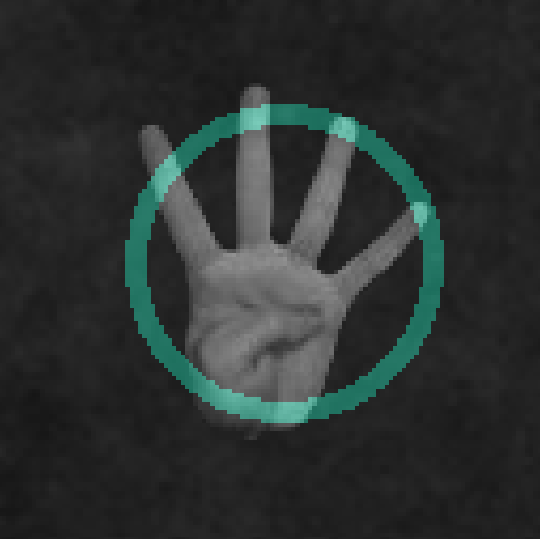}}
\caption{Circle applied on image}
\label{circle_hand}
\end{figure}

\begin{figure}[htbp]
\centerline{\includegraphics[width=50mm,scale=0.5]{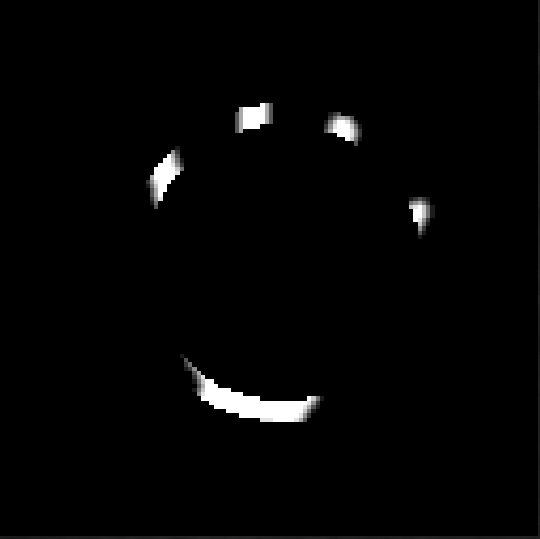}}
\caption{Circle applied as a mask on image}
\label{circle_fingers}
\end{figure}

Using a tool from \cite{cv2}, we can find the contours of the mask (Fig. \ref{mask_hand}), apply the Convex Hull function to remove the concavity between the finger, and obtain a geometric shape from it, as can be found on Fig. \ref{contour_hand}. With the coordinates of this geometric shape, we can obtain the hand's centroid and size, to be used as parameters to draw a circle that covers all the fingers and the wrist of the hand, as we can see on Fig. \ref{circle_hand}. Applying this circle as a mask and finding its contours, as can be found on Fig. \ref{circle_fingers}, we can obtain only the parts which the fingers and wrist touch the circle.

\subsubsection{Counting Heuristic}
To develop a heuristic to count the number of fingers an image may have, first we use the information of each contour's size found on Fig. \ref{circle_fingers} to identify if there is a wrist, using a conditional to remove a contour with a higher arbitrary maximum size. As final, we count the remaining contours and the result will be the number of fingers on this image.

\subsection{Convolutional Neural Network}

To introduce the Convolutional Neural Network model it is important to understand better how some of its parts work. This is a rough overview, and a survey with a thorough explanation can be found on \cite{b2}.

\begin{itemize}
\item \textbf{Activation} In a Neural Network, we define how we will output the values from one layer through the application of an activation function. A neural network can have different activation functions in different parts of its network, as explained by the fact that an activation function can have a high impact on network learning. An in-depth explanation of activation functions and their use case can be found on \cite{b4}.

An example of possible activation functions can be found on Fig. \ref{activation_functions}.

\begin{figure}[htbp]
\centerline{\includegraphics[width=90mm,scale=0.5]{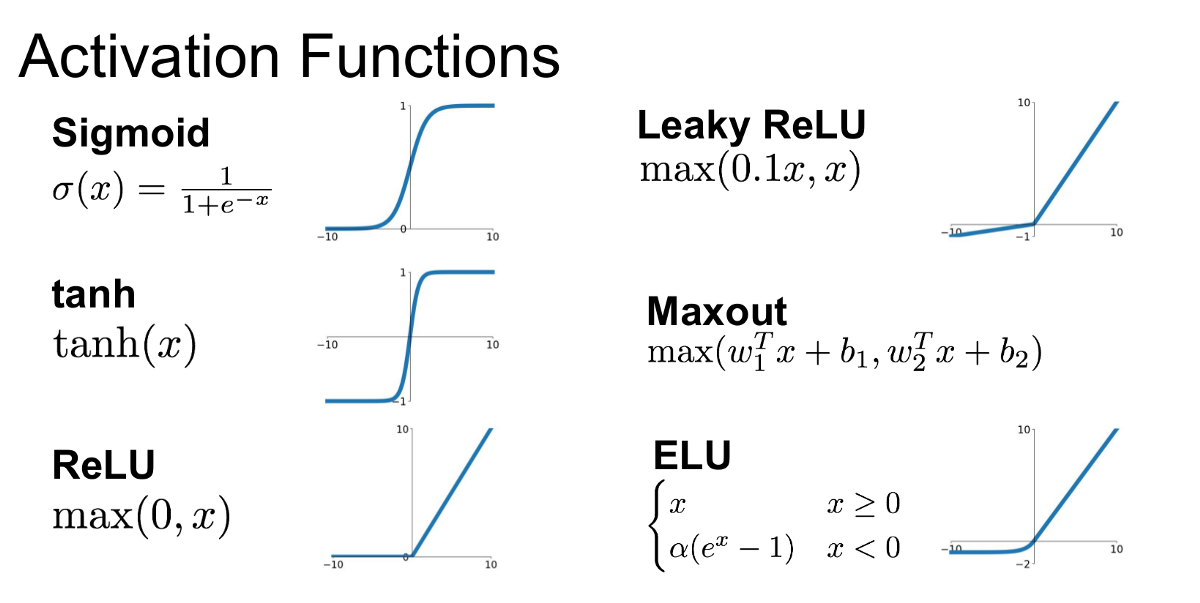}}
\caption{Possible activation functions \cite{b4}}
\label{activation_functions}
\end{figure}

\item \textbf{Batch Normalization Layer} Batch Normalization layers were introduced on \cite{b5} as a means to avoid Internal Covariate Shift, which, according to \cite{b5}, can be defined ``[...] as the change in the distribution of network activations due to the change in network parameters during training.'', i.e., the fact that when our data passes from a layer to the following one the data on it can vary its range creating numerical instability. Intuitively, what this layer does is to re-scale and re-center our layer values in an acceptable (and learnable) range.

\item \textbf{Max-Pooling Layer}

A max-pooling layer is a non-learnable layer that decreases our data shape/size. A simple and intuitive explanation can be found on Fig. \ref{max_pooling}. This layer has a kernel with size $k \in \mathbb{R}^n$ which is iterated through the input data, reducing the input dimensionality. For example, for a matrix $M \in R^{nXm} $, if we applied a $(i, j)$ kernel-sized max-pooling layer to it we would end up with a new matrix $N \in R^{\lfloor n/i\rfloor X \lfloor m/j\rfloor}$.

\begin{figure}[htbp]
\centerline{\includegraphics[width=70mm,scale=0.5]{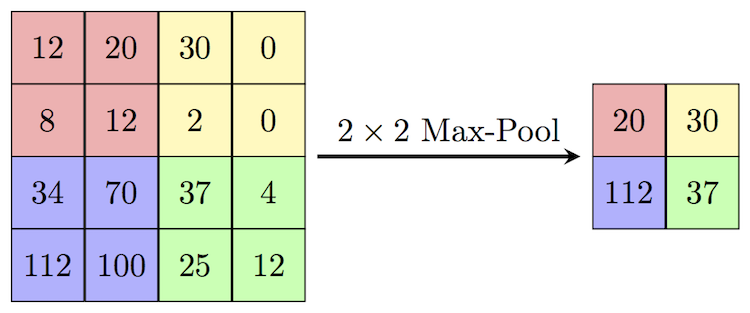}}
\caption{Example of a 2D MaxPooling application using a 2x2 kernel \cite{b3}}
\label{max_pooling}
\end{figure}

\item \textbf{Dropout Layer}

Proposed on \cite{b6}, dropout layers are used to avoid overfitting. Randomly dropping some of the previous layer values, we avoid the network completely memorizing the input since it needs to keep track of the classification however values are randomly dropped. Note that we do not always the same values, and in fact, we drop every feature with probability $p$ using samples from a Bernoulli distribution.

\end{itemize}

\subsection{The model - FINNger}

The model, from now on referred to by its name FINNger, was developed in Python using PyTorch \cite{pytorch_paper}\cite{pytorch_website}. We also relied heavily on cv2\cite{cv2} and numpy\cite{numpy} to parse and manipulate the images.

We trained several different models before settling with the one described below. In the beginning, we tried detecting both the quantity of the numbers and the hand (left or right) on the image, given the fact that the dataset had this information. After, we ended up dropping this because we didn't use this information for anything in our work.

\begin{figure*}
\centerline{\includegraphics[width=150mm,scale=0.8]{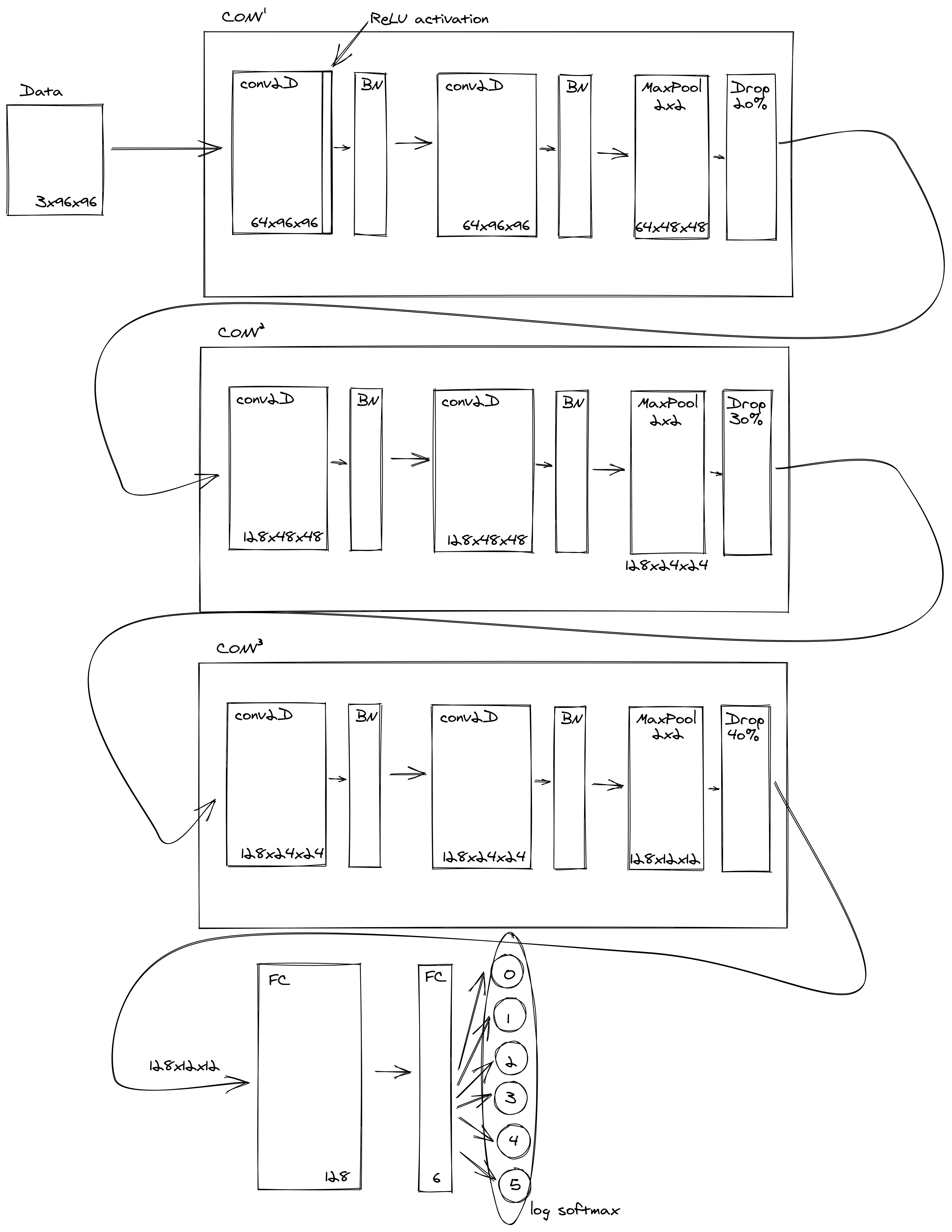}}
\caption{Our model}
\label{model}
\end{figure*}

Remember that batch normalization and dropout layers do not change the input shape, outputting the same shape. Also, every convolutional layer uses a $(4,4)$ kernel size. It is built in the form of the following building blocks, remembering that the input has shape $(3,96,96)$:

\begin{enumerate}
    \item $C^1$: Convolutional block 1, composed of the following layers:
    \begin{enumerate}
        \item $C^1 c^1$: 2D Convolutional layer outputting a $(64,96,96)$ matrix, with ReLU activation
        \item $C^1 b^1$: Batch normalization layer
        \item $C^1 c^2$: 2D Convolutional layer outputting a $(64,96,96)$ matrix, with ReLU activation
        \item $C^1 b^2$: Batch normalization layer
        \item $C^1 mp_{2X2}$: 2D MaxPooling layer with a $(2, 2)$ kernel, outputting a $(64, 48, 48)$ matrix
        \item $C^1 d_{0.2}$: Dropout layer with $p=0.2$, this is, setting, randomly, 20\% of the input values to 0
    \end{enumerate}
    
    \item $C^2$: Convolutional block 2, composed of the following layers:
    \begin{enumerate}
        \item $C^2 c^1$: 2D Convolutional layer outputting a $(128,48,48)$ matrix, with ReLU activation
        \item $C^2 b^1$: Batch normalization layer
        \item $C^2 c^2$: 2D Convolutional layer outputting a $(128,48,48)$ matrix, with ReLU activation
        \item $C^2 b^2$: Batch normalization layer
        \item $C^2 mp_{2X2}$: 2D MaxPooling layer with a $(2, 2)$ kernel, outputting a $(128, 24, 24)$ matrix
        \item $C^2 d_{0.3}$: Dropout layer with $p=0.3$, this is, setting, randomly, 30\% of the input values to 0
    \end{enumerate}
    
    \item $C^3$: Convolutional block 3, composed of the following layers:
    \begin{enumerate}
        \item $C^3 c^1$: 2D Convolutional layer outputting a $(128,24,24)$ matrix, with ReLU activation
        \item $C^3 b^1$: Batch normalization layer
        \item $C^3 c^2$: 2D Convolutional layer outputting a $(128,24,24)$ matrix, with ReLU activation
        \item $C^3 b^2$: Batch normalization layer
        \item $C^3 mp_{2X2}$: 2D MaxPooling layer with a $(2, 2)$ kernel, outputting a $(128, 12, 12)$ matrix
        \item $C^3 d_{0.4}$: Dropout layer with $p=0.4$, this is, setting, randomly, 40\% of the input values to 0
    \end{enumerate}
    
    \item $FC^1$: Fully connected layer, which flattens the previous $\mathbb{R}^3$ matrix to a one-dimensional vector when taking it as input. It then outputs a $(128)$ sized vector.
    \item $FC^1$: Fully connected layer, responsible for outputting the classification, and, therefore, outputting a $(6)$-shaped vector. In the end we apply a "log softmax" function to the output, as shown in Eq. \ref{log_softmax}:
    
    \begin{equation*}
        \text{LogSoftmax}(x_i) = \log \left( \frac{\exp{x_i}}{\sum_j \exp{x_j}}  \right)
        \label{log_softmax}
    \end{equation*}
    
A schema representing the full model architecture can be found on Fig. \ref{model} where the displayed numbers represent the output shape of that layer.
    
\end{enumerate}

\section{Results}
\subsection{Edge Detection}
\begin{figure}[htbp]
\centerline{\includegraphics[width=95mm]{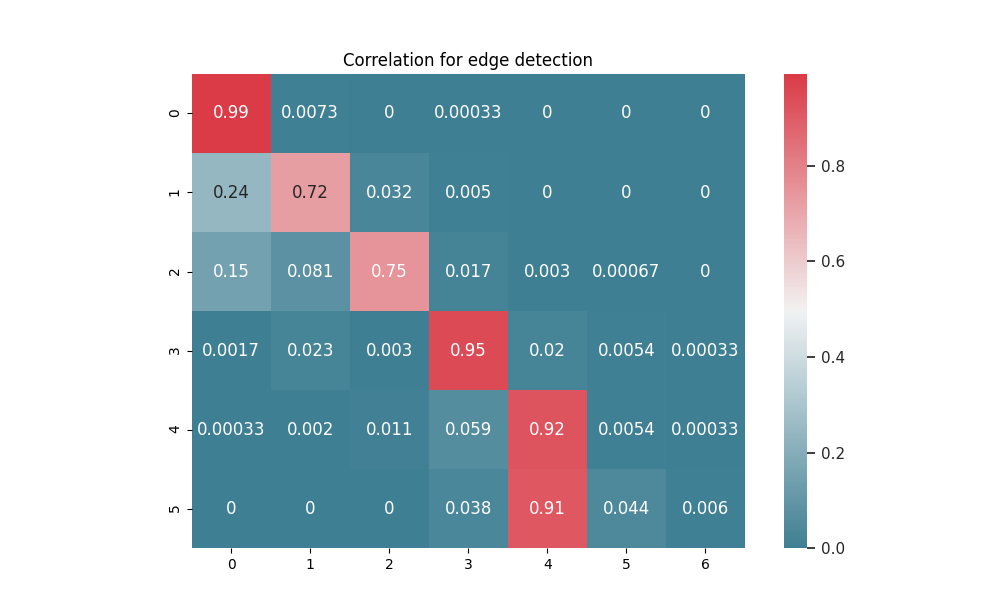}}
\caption{Correlation between the expected values (lines) and the heuristic output (columns)}
\label{edge_correlation}
\end{figure}

We evaluated all the images from the koryakinp/fingers dataset to analyze how well the Edge Detection performed. We were able to achieve the success accuracy of 72.4\% using the testing images and 72.9\% using the training images, with a standard deviation of 1.15 fingers on both. 
We used these results to create a correlation graph, as you can see on Fig. \ref{edge_correlation}, and we could notice that the Edge Detection had most of its errors when evaluating images with 5 fingers, where the results took were evaluated missing 1 finger on the image. This happened mostly because the process to differentiate the fingers couldn't distinguish between touching fingers. 
We can notice that there were some evaluations to 6 fingers, where the conditional to remove the wrist didn't work as expected, where the wrist's size wasn't big enough to be caught in the condition. This behavior occurred in the results with a value of 0.08\% using the testing images and 0.11\% using the training images.
The Edge Detection applied to the custom dataset couldn't work satisfactorily, there was almost 30\% of accuracy. However, in general, all the evaluations were false positives, when the evaluation results were unintentionally successful. We analyzed the evaluation process and noticed that the HSV conversion and masking to separate the hand from the background as we used on Fig. \ref{HSV_hand} and Fig. \ref{contour_hand}, didn't work as expected, separating parts of the background because the environment wasn't controlled as the koryakinp/fingers dataset.

\subsection{FINNger}
On early attempts we used \cite{b7}'s dataset, but it ended up not having good results when facing real-life examples, even though it achieved $92\%$ accuracy on the validation dataset after only $10$ epochs. We also tried using our custom dataset mixed with \cite{b7} but it ended up ignoring our dataset, and training/learning only \cite{b7} characteristics. The possible problem with the dataset is the fact that the images are more illuminated than the background (in real life this is probably the opposite), which highlights the fact that the network worked at night when we had a dark background and an indirectly lightened hand.

We then changed our approach to using only our custom dataset, implementing data augmentation to fix the fact that the dataset is small when considering the size of the datasets that deep learning usually needs. Our data augmentation process was fairly simple, creating random translations and horizontal flip of the image. We did not implement rotation data augmentation and this can be seen in the empirical results when trying to use the resulting network, noticing that if we tilt the hand a little bit, the accuracy decreases drastically.

We trained our FINNger neural network for 100 epochs with \textit{Adam}\cite{adam} optimizer. As hyperparameters, we chose $0.0003$ for the learning rate and $0.0001$ for the weight decay. We didn't do any exploration with the hyperparameters since these values were the first tested ones (based on previous works neural networks) and they worked pretty well.
 
\begin{figure}[htbp]
\centerline{\includegraphics[width=95mm]{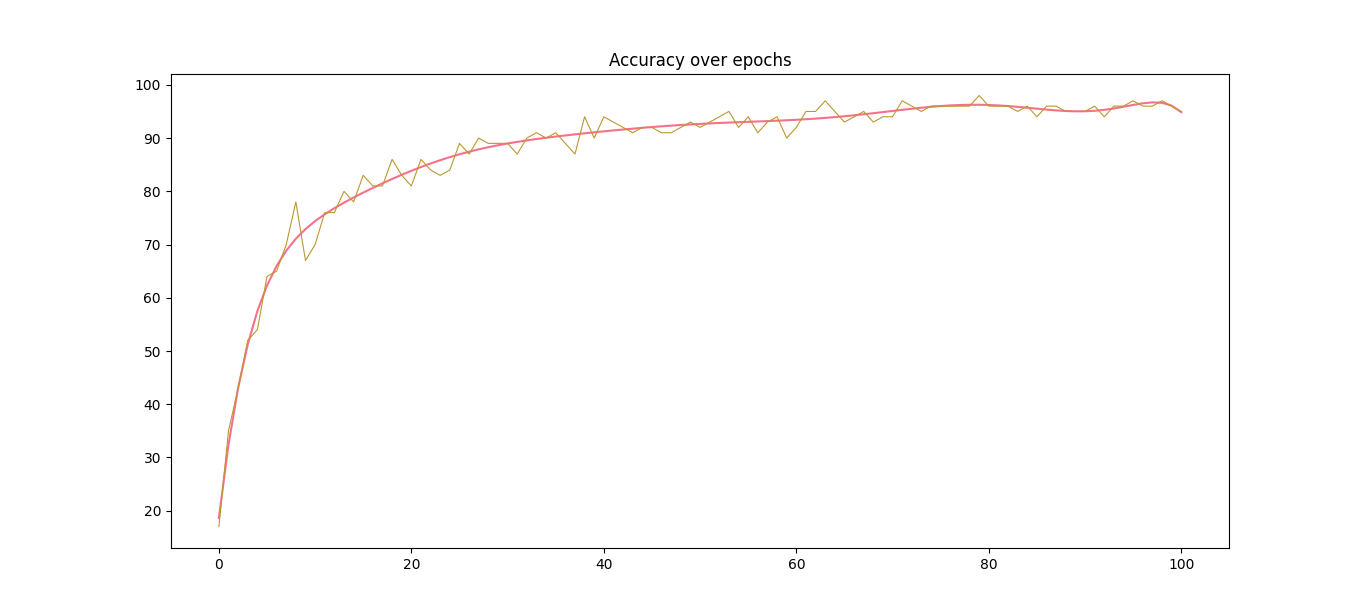}}
\caption{Model accuracy over the 100 epochs where the brown line depicts the exact accuracy and the orange line a smoothed polynomial}
\label{finnger_accuracy}
\end{figure}

In our validation dataset, we achieve a maximum $98\%$ accuracy as depicted on Fig. \ref{finnger_accuracy} after circa 75 epochs, and capping around it oscillating back and forth to $95\%$.

\begin{figure}[htbp]
\centerline{\includegraphics[width=95mm]{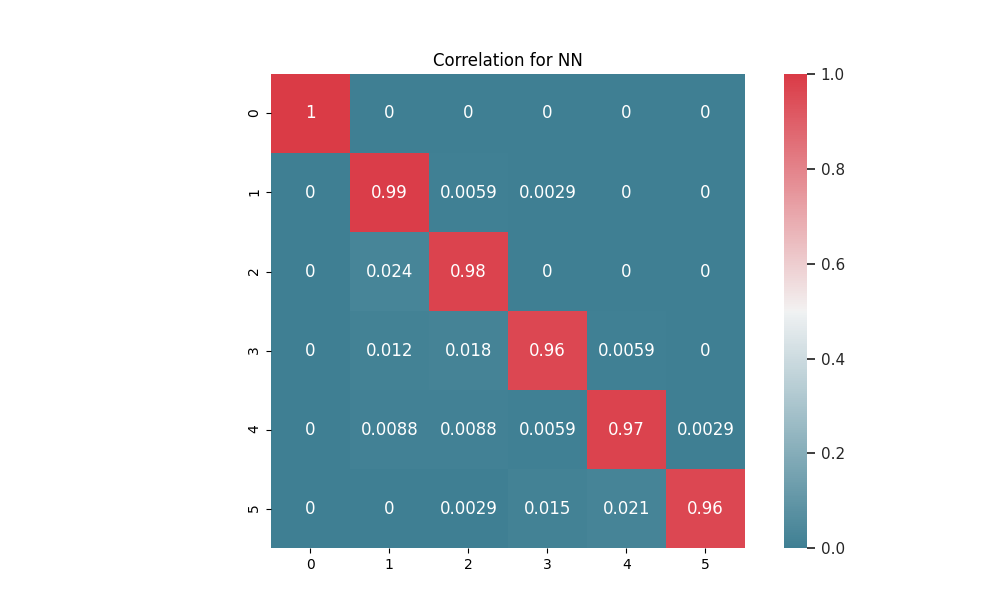}}
\caption{Correlation between the expected values (lines) and the network output (columns)}
\label{finnger_covariance}
\end{figure}

Plotting a correlation matrix, as seen on Fig. \ref{finnger_covariance} we can see that we have a high correlation, and when missing the correct value we rarely overestimate, tending to get the results wrong saying we have fewer numbers than we have. Important to notice that we will never say we have more than 5 fingers (like we did on the ``Edge Detection'' approach) because the neural network can only classify the input image in 6 categories, namely $[0, 1, 2, 3, 4, 5]$.

When tested against real children's hands, we don't achieve good results. The network tends to detect $0$ fingers, except when the children have the hand fully open. This is probably related to the fact that the proportion of children's hands are different than from an adult one, with the fingers being a lot smaller in comparison. We hypothesize however that if we trained our neural network using children's hands' images we could achieve an acceptable result.

\begin{figure}[htbp]
\centerline{\includegraphics[width=70mm]{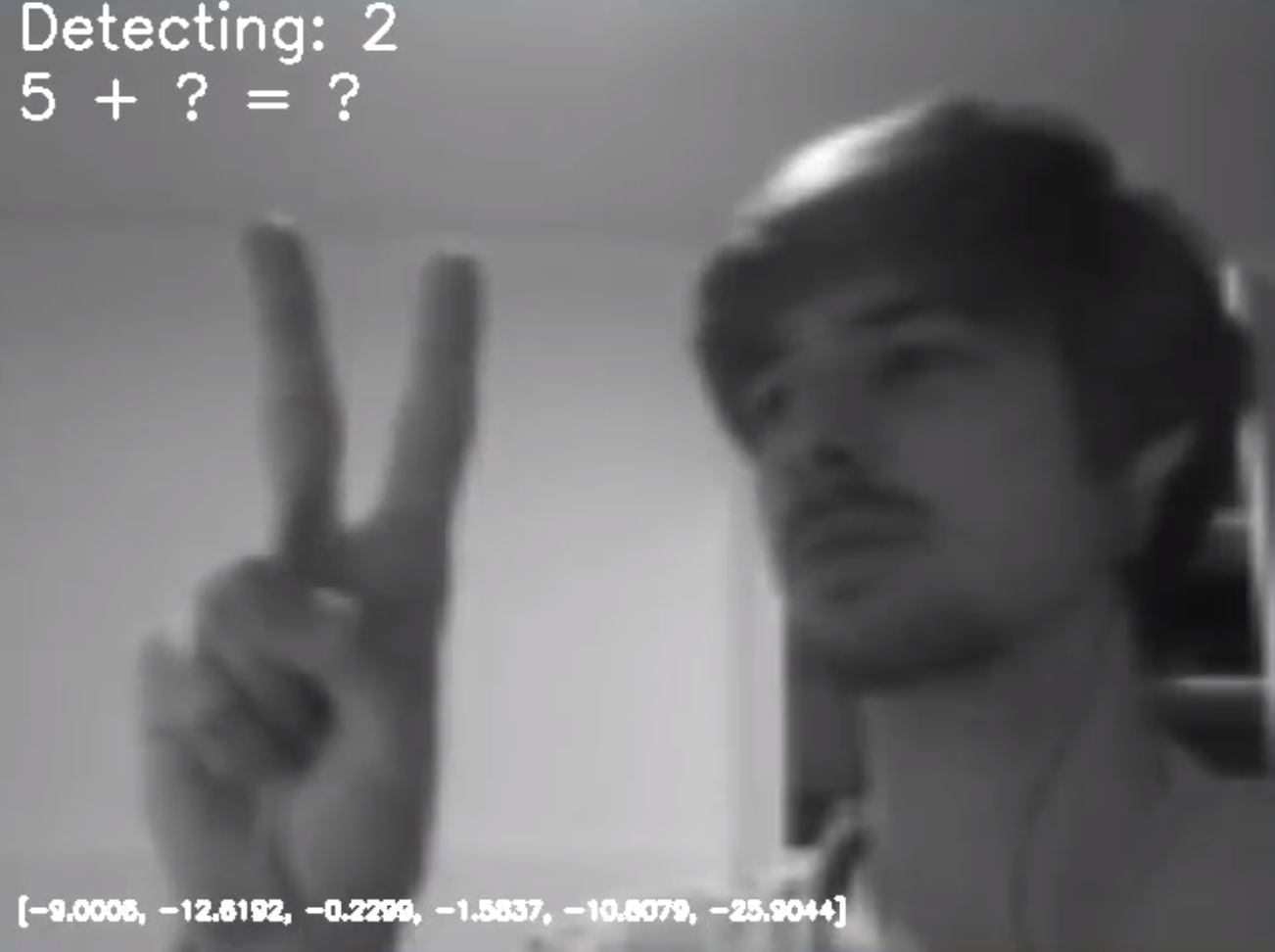}}
\caption{UI from our application}
\label{app_ui}
\end{figure}

Also, we developed a basic UI, see Fig. \ref{app_ui}, where we can check the UI result and do the calculations as we proposed so that it could be used by children to learn basic arithmetic. On the top, we have the value that the neural network currently thinks it is being shown to it. Below it, we display the calculation: to compute it you press enter on the terminal when you want to "capture" the result. On the bottom, we are displaying, for debug purposes, the values outputted by the network.

You can verify on the example (Fig. \ref{app_ui}) that it is correctly detecting a $2$. Checking the network output we can see that it has great confidence that it is a $2$, although it also thinks it might be a $3$ from the value in the fourth vector position, which makes sense since $3$ has the more similar "hand format" to $2$.

The code for the model (training and evaluation), as well as the dataset creation application can be found on \cite{github_code}.

\section{Drawbacks}
In the process to evaluate hands with different colors or different background, there were many cases where the Edge Detection couldn't distinguish a hand successfully, these are related to the wide range of colors the background and the hand can variate, making it difficult to find the right specific range of colors for a hand. And because the background can have the same colors as a hand, this approach has some limitation to not mix the background within the hand's Edge Detection.

The Convolutional Neural Network couldn't be trained with a diverse of scenarios enough to learn about hands with different colors or sizes such as child's hands. This limitation restricted the accuracy when applying the CNN to a wide range of scenarios. When evaluating with children, because of the small size of the hands, the results were mostly evaluated to zero fingers (or closed hand). We also noticed that what the CNN was to distinguish lines on the images, which made it find fingers on objects with shapes as straight lines, such as furnishings, bookshelves, or any other objects on the scenario. We believe these errors could reduce using better quality images with a higher range of colors (instead of training it with B\&W images). 

\section{Future Work}
To build onto this work, we could work on the following:

\begin{itemize}
    \item Improve the finger counting heuristic on the edge detection algorithm, to better use the convex hull, possibly counting the number of outstanding edges instead of using the ``circle'' technique;
    \item Create a bigger and more comprehensive dataset using different illumination and location settings. Also, actually using children hands so that the neural network can correctly infer there fingers (learning that fingers can be short);
    \item Create a syntetic dataset using \cite{b7}'s hands and adding custom random backgrounds, changing the illumination, rotation and translation generating, possibly, an infinitely-sized dataset;
    \item Create a mobile application which uses the trained model, so that it can be more easily acessed.
\end{itemize}

\section{Conclusion}
We showed that we can have basic and satisfactory results using a good edge detection algorithm if we have strict and controlled position and illumination settings.

For the neural network, we arrived at the conclusion that we had as the hypothesis in the beginning: it would work for similar images to the dataset, but probably not achieve optimal results with children hands. We hypothesize that the only gap to be filled to achieve a perfect application is a better dataset. Using any of the combinations of the techniques proposed in the ``Future Work'' section would also improve our work and make it better to be used in real life.

\end{document}